\documentclass[letter, 10 pt, conference]{ieeeconf}

\usepackage{fixfoot}
\usepackage[utf8]{inputenc}
\usepackage{graphicx,dblfloatfix,caption,afterpage}
\usepackage{amsmath}
\usepackage{mathtools}
\usepackage{mathtools, cuted}
\usepackage{lipsum, color}
\usepackage{float}
\usepackage{cuted}
\usepackage{subfloat}
\usepackage{subfigure}
\usepackage{tabularx}
\usepackage{color}
\usepackage{multirow}
\usepackage{array}
\usepackage{tensor}
\usepackage{scalerel,stackengine}
\stackMath
\newcommand\reallywidehat[1]{%
\savestack{\tmpbox}{\stretchto{%
  \scaleto{%
    \scalerel*[\widthof{\ensuremath{#1}}]{\kern-.6pt\bigwedge\kern-.6pt}%
    {\rule[-\textheight/2]{1ex}{\textheight}}
  }{\textheight}%
}{0.5ex}}%
\stackon[1pt]{#1}{\tmpbox}%
}
\input epsf
\newcolumntype{C}[1]{>{\centering\arraybackslash}m{#1}}

\makeatletter
\newcommand*{\rom}[1]{\expandafter\@slowromancap\romannumeral #1@}
\makeatother

\IEEEoverridecommandlockouts 

\title{Estimation of Trocar and Tool Interaction Forces on the da Vinci Research Kit with Two-Step Deep Learning} 
\author{Jie Ying Wu$^{1,*}$, Nural Yilmaz$^{2,*}$, Peter Kazanzides$^{1,\dagger}$, Ugur Tumerdem$^{2,\dagger}$
\thanks{$^{1}$Dept. of Computer Science, Johns Hopkins University, Baltimore, MD 21218, USA {\tt\small (email: \{jieying, pkaz\}@jhu.edu)}}%
\thanks{$^{2}$Dept. of Mechanical Engineering, Marmara University, Istanbul, Turkey {\tt\small (email: nural.yilmaz@marun.edu.tr, ugur.tumerdem@marmara.edu.tr)}}%
\thanks{$^{*}$Joint first authorship, $^{\dagger}$Joint senior authorship.}
}

\begin{document}

\maketitle

\begin{abstract}
Measurement of environment interaction forces during robotic minimally-invasive surgery would enable haptic feedback to the surgeon, thereby solving one long-standing limitation. Estimating this force from existing sensor data avoids the challenge of retrofitting systems with force sensors, but is difficult due to mechanical effects such as friction and compliance in the robot mechanism. We have previously shown that neural networks can be trained to estimate the internal robot joint torques, thereby enabling estimation of external forces. In this work, we extend the method to estimate external Cartesian forces and torques, and also present a two-step approach to adapt to the specific surgical setup by compensating for forces due to the interactions between the instrument shaft and cannula seal and between the trocar and patient body. Experiments show that this approach provides estimates of external forces and torques within a mean root-mean-square error (RMSE) of 2\,N and 0.08\,Nm, respectively. Furthermore, the two-step approach can add as little as 5 minutes to the surgery setup time, with about 4 minutes to collect intraoperative training data and 1 minute to train the second-step network.

\end{abstract}

\section {Introduction}

Robotic surgery has enabled better surgeon ergonomics and patient care. Yet this comes at the cost of increased distance between surgeon and patient, which also results in a lack of  direct haptic feedback to the surgeon \cite{Okamura2004}. Although surgeons can learn to use visual feedback to adjust for the lack of haptic feedback, restoring it can make telerobotic surgery easier to learn, more intuitive, and precise. 

\begin{figure}[!t]
\begin{tabular}{cc}
\centering
\includegraphics[height=0.22\textheight]{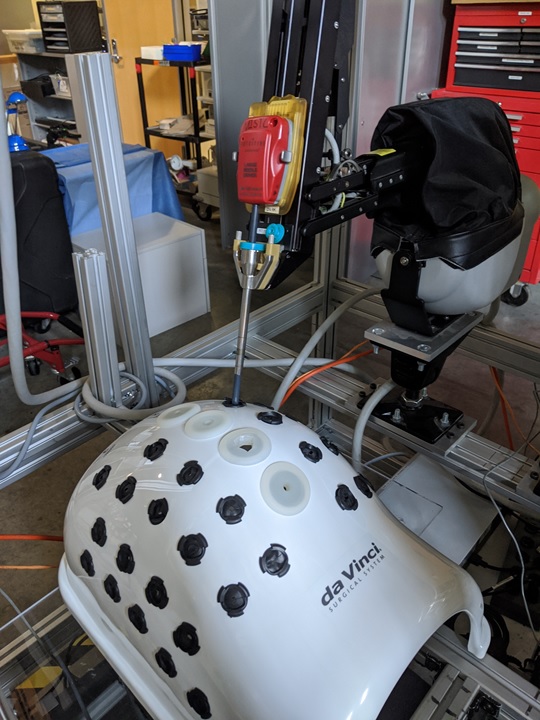}	&
\includegraphics[height=0.22\textheight]{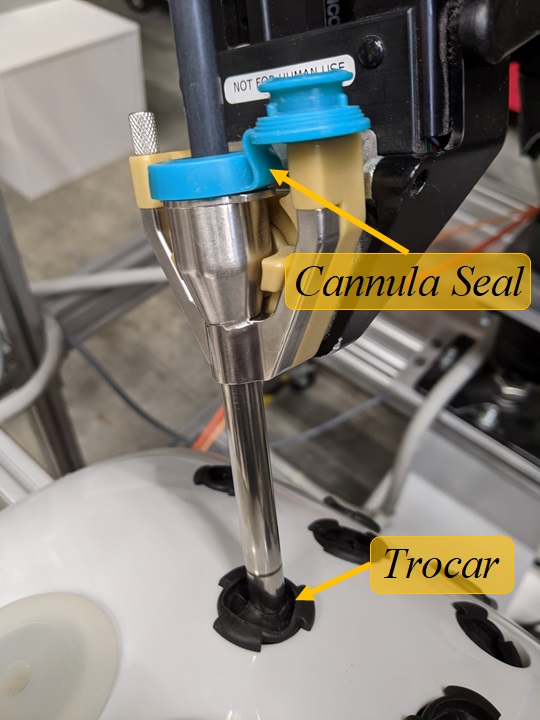} \\
\end{tabular}
\caption{Left: Trocar with cannula seal; Right: closeup view.}
\label{fig:trocar}
\end{figure}

Adding force sensing capability to surgical robots has been an active research subject and multiple alternative
solutions have been suggested in the literature. In \cite{Madhani}, force estimation has been performed with joint current/torque measurements of the Black Falcon surgical robot, but it has been reported that inertial forces of the robot are dominant in torque measurements, especially in free motion. When reflected to the operator they can be annoying. Fig.~\ref{fig:trocar} shows potential contributors to free motion inertial forces as they interact with the instrument along its length rather than at the tip. 

One way to solve this problem is to make use of force sensors in the patient side manipulators. In \cite{berkelman2003miniature}, three axis, and in \cite{seibold2005prototype},  six axis miniature force sensors have been manufactured and attached to the tips of specially developed surgical instruments. Due to the difficulties in manufacturing such sensors, other researchers have attached flexible capacitive \cite{kim2015force}, strain gauge based \cite{Pena2018}, or fiber-optics based \cite{peirs2004micro} sensors on surgical grippers. However, the force/torque sensing degrees of freedom are limited in these approaches. The common drawbacks of all intra-corporeal sensor based approaches is the difficulty in manufacturing accurate and robust multi-DOF sensors that can be miniaturized, as well as the sterilization requirements after each operation \cite{willaert2013design}. 

To overcome these problems, some researchers have proposed the use of force sensors placed at the noninvasive (extra-corporeal) part of the robot. It is possible to estimate tip forces through extra-corporeal sensors, but since the robot interacts with the patient body at the trocar these forces would also be registered by the sensor. In order to filter these forces, the ``overcoat'' method was proposed in \cite{shimachi2003measurement} by introducing a second trocar sensor. This method has been further developed in \cite{willaert2013design} and \cite{schwalb2017force}.  Since these methods require bulky force sensors placed at the trocar, \cite{fontanelli2017novel} considers the use of fiber optic sensors in the trocar.

These methods can provide accurate estimates of forces in the Cartesian XYZ directions of the tool tip. However, the wrist bending torques cannot be estimated/measured. Furthermore, the placement of force/torque sensors in the trocar may not always be clinically feasible. 

Another approach to eliminate the above problems is to estimate forces based on robot dynamics identification. In order to filter out the robot dynamic forces interfering with the external force estimation, an accurate dynamic identification of the surgical manipulator is typically required. Friction compensation has been performed on a da Vinci system in \cite{mahvash2008force}, cable tension estimation with dynamic identification has been performed on the Raven II system in \cite{haghighipanah2017utilizing} and inverse dynamics estimation has been performed on a modified da Vinci gripper system in \cite{lee2010reaction}. These papers have reported force estimation in 1 DOF. In \cite{sang2017external}\cite{pique2019dynamic}, more comprehensive explicit model-based dynamic identification has been used in the estimation of external forces on the da Vinci Research Kit (dVRK). In \cite{sang2017external}, 3 DOF Cartesian tip forces have been estimated in a pseudo-static configuration and in \cite{pique2019dynamic} 6 DOF tip forces and wrenches have been estimated in a more dynamic scenario. Although these model-based approaches have good performance, they do not take patient-robot interactions into consideration. Since all possible configurations during surgery have to be considered in the explicit models, this would make the model-based approach difficult to implement in clinical scenarios. 

With the recent advances in machine learning methods, some researchers have applied deep learning to force estimation in robotic surgery. In \cite{Oneill}\cite{guo2019grip}, the authors have trained deep neural networks to estimate gripper forces on 1 DOF instruments, by using joint and force sensor measurements in training. In \cite{yu2020microinstrument}\cite{tran2020}, a similar approach has been used for Cartesian force estimation in three axes. However, these approaches require the use of a force sensor for ground truth and do not take robot-patient interactions into account in the model training. The use of a force sensor for training in a patient would be difficult, limiting the applications of these approaches in a clinical scenario. 

In our previous works \cite{yilmaz2018external}\cite{yilmaz20196}\cite{yilmaz2020neural}, a dynamic identification and external force estimation approach similar to \cite{sang2017external} and \cite{pique2019dynamic} was presented on different surgical systems, but instead of using explicit mathematical models to identify and filter out dynamic forces, neural networks were used. 

In theory, training of neural networks can be performed even in the surgical theatre where various robot configurations and patient/robot interactions are possible, as the black box nature of the neural networks eliminates the need for an explicit interaction model. However, in that case, the main issue becomes the feasibility of training in a clinical setting, which is directly related to the required amount of data collection by the surgeon (labor cost) and how long it takes for the training to be completed (computational cost). While these issues have not been addressed in previous works, they form the basis of our proposal in this paper. 
\subsection{Contributions}

In this paper, we extend our use of neural networks to identify and eliminate trocar interaction forces. As the trocar interaction forces are patient-specific, we use two-step deep learning techniques to make this identification feasible in a clinical scenario. The proposed technique is completely self-supervised to not require additional setup in the clinic. In addition, we extend our prior work to estimate Cartesian torques in addition to forces.

\section{Methods}
To account for trocar interaction forces, we propose a two-step learning scheme. The first network, described in Section \ref{sec:step_one}, follows our previous work in estimating free space dynamics \cite{yilmaz2020neural}. It is trained extensively for robot joint torque identification in the robot workspace. It is followed by another network that learns patient and setup specific effects. The second network is designed to be trained with a subset of the workspace that is relevant to the procedure after the robot has been docked and the instruments have been placed through the port. Section \ref{sec:step_two} presents two variants of the second-step network. 

\begin{figure}[!t]
\centering
\includegraphics[width=0.5\textwidth]{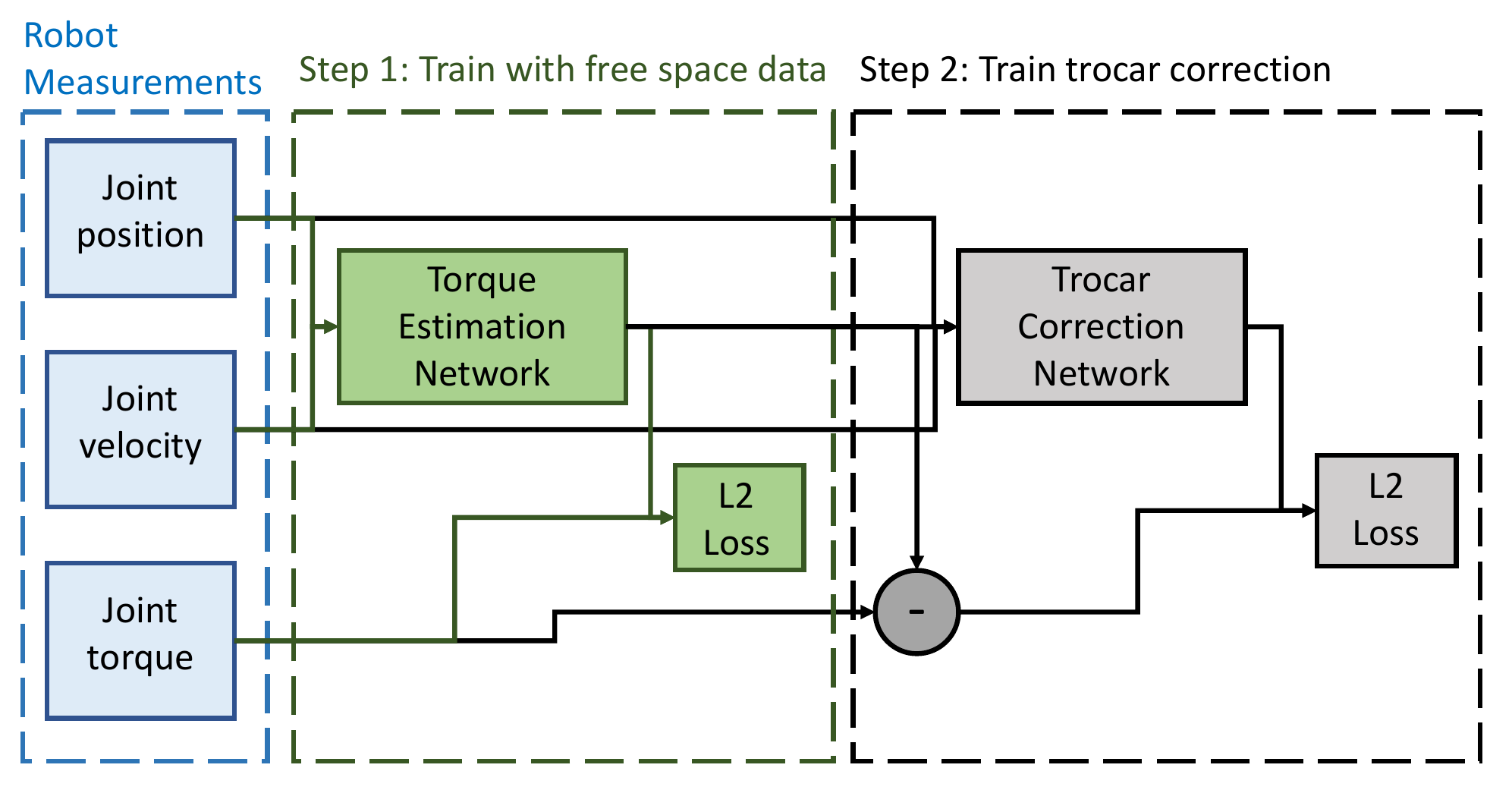}
\caption{Block diagram of two step correction network (\textbf{Corr})}
\label{fig:block_diagram}
\end{figure}

\subsection{Step 1: Joint Torque Identification Network}
\label{sec:step_one}
Following the comparison in \cite{hwang2020efficiently}, we adopt an LSTM as our base network for torque identification without the trocar. The inputs to the network are the positions and velocities of all the joints. Joint positions are read from encoders and joint velocities are calculated according to \cite{Wu2018}. The network architecture consists of an LSTM layer with 128 hidden dimensions followed by a fully connected layer to the output dimension. Since the third joint has hysteresis (due to its interaction with the cannula seal) that is more complicated to model, we add a hidden layer with 128 nodes after the LSTM and before the output layer for this joint. We use ReLU activation and drop out after the 128-node hidden layer to prevent over-fitting. 

\subsection{Step 2: Compensation for Trocar Interaction Forces}
\label{sec:step_two}

We insert the da Vinci Patient Side Manipulator (PSM) instrument inside an abdominal phantom (Intuitive Surgical Inc., Sunnyvale CA, USA) such that its remote center of motion is at the insertion port, as shown in Fig.~\ref{fig:trocar}. We implemented and evaluated two methods for compensating for the trocar interaction forces, as described in the following two subsections. With these methods, it is possible to improve the joint torque identification network by taking only a small amount of data collected in a specific patient setup. 

\subsubsection{Correction network (Fig.~\ref{fig:block_diagram})}
\label{sec:corr}

Since LSTMs are good at capturing time series data but are generally slower to train, we consider a feed-forward network for learning the patient-specific trocar correction. The network aims to learn the residual error between the network trained without the trocar and the measured torque during free space motion. To capture some time-series information without the heavy computation cost of LSTMs, we use a window of previous position and velocity measurements of all joints and the predicted free space torques as input. The network has one layer of 256 hidden nodes and one to two outputs. ReLU activation is used between the hidden and output layer. The output of the network is added to the free space network prediction as a correction factor. 

\begin{figure}[!t]
\centering
\includegraphics[width=0.48\textwidth]{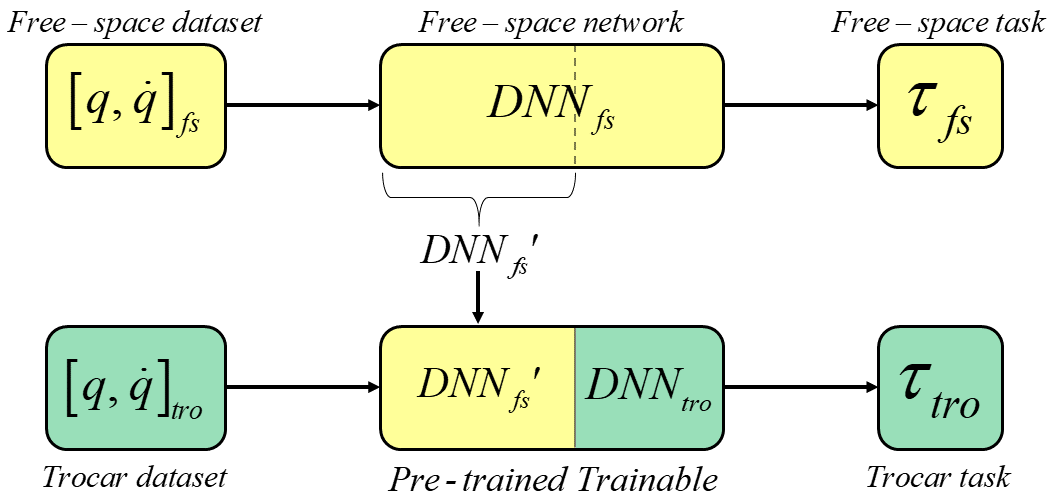}
\caption{Block diagram of transfer learning (\textbf{Xfer}) method (fs: free-space, tro: trocar, DNN: Deep neural network)}
\label{fig:transLearn}
\end{figure}

\subsubsection{Transfer Learning Method}
As an alternative, we also consider a transfer learning approach. We truncate the free space network after the LSTM layer and append another LSTM and fully connected layers. During training with trocar data, we freeze the free space layers. As can be seen in Fig.~\ref{fig:transLearn}, the same free space network can be transferred for use in a different task, and selected layers of it can be retrained to adapt to the new task, providing improvements in training times. In the first LSTM layer transferred from the pre-trained network, the learning rates are set to be zero while the appended layers are trained with the new data. 

\section{Experimental Setup}

\subsection{Free Space Torque Identification Network Training}
Similar to the setup in \cite{yilmaz2020neural}, we move the robot instrument in free space. We collect about an hour of data moving in free space without any contact, and with and without the cannula seal. We use 80\% of the data as our training set and 10\% each for the test and validation sets. Extending our previous work, which only explored a limited workspace, we collected as full a workspace as possible and the workspace analysis is shown in Fig.~\ref{fig:workspaces}. 

We implement the free space network described in Sec.~\ref{sec:step_one} in Matlab and train two versions, one with the cannula seal and one without. We use the Adam optimizer with initial learning rate set to 0.001. We decay the optimizer every 125 epochs by a factor of 0.5. We train for 1000 epochs. To balance the computation cost of training networks for each joint vs. mixing losses that are different, we train 1 network for joints 1 and 2, and 1 network for joints 5 and 6. Joints 3 and 4 have respectively a much larger and smaller range of torques than the other joints and therefore are trained as separate networks. 

\subsection{Trocar and Cannula Seal}

To learn the trocar effects, we collect about 20 min of interactions inside the cannula and split it as 80\%/10\%/10\% into train, validation, and test sets. We insert the cannula into a trocar hole in the abdominal phantom and place a cannula seal on top. Then, we attach the cannula to the dVRK and insert the instrument through the cannula. 

For each free space network, we train a set of correction networks and a set of transfer learning correction layers. Since the free space network trained with the cannula seal matches the test conditions better, we expect that it will perform better than the free space network trained without the cannula seal. However, correcting the latter free-space network enables us to estimate how well the correction schemes can generalize to test setups that are extremely different from training. The correction network is trained for 400 epochs with an initial learning rate of 0.001. We use 0.01 L2 regularization on the weights to ensure it does not overfit to small datasets.  The window size was set to be 10 empirically through grid search. 

For the transfer learning approach, the learning rate is set to 10 to reduce the number of epochs for which the LSTM must be trained. Both approaches use the Adam optimizer. Our configurations are summarized in Table~\ref{table:terms}, where the \textbf{Base} configuration is consistent with our prior work \cite{yilmaz2020neural}.

\begin{table}[h]
\centering
\caption{Tested Network Configurations}
\label{table:terms}
\begin{tabular}{|p{0.8cm}|p{6.5cm}|}
\hline
\textbf{Troc} & Trained directly from trocar using the structure described in \ref{sec:step_one}. \\ \hline
\textbf{Base} & Trained on data collected without the trocar and without the cannula seal as described in \ref{sec:step_one} \\ \hline
\textbf{Seal} & Trained on data collected without the trocar but with a cannula seal \\ \hline
\textbf{Corr} & Add a correction trained from trocar data using second network \\ \hline
\textbf{Xfer} & Add a correction trained from trocar data using transfer learning \\ \hline 
\end{tabular}
\end{table}

\begin{figure*}[h]
\begin{tabular}{cc}
\centering
\includegraphics[width=0.49\textwidth]{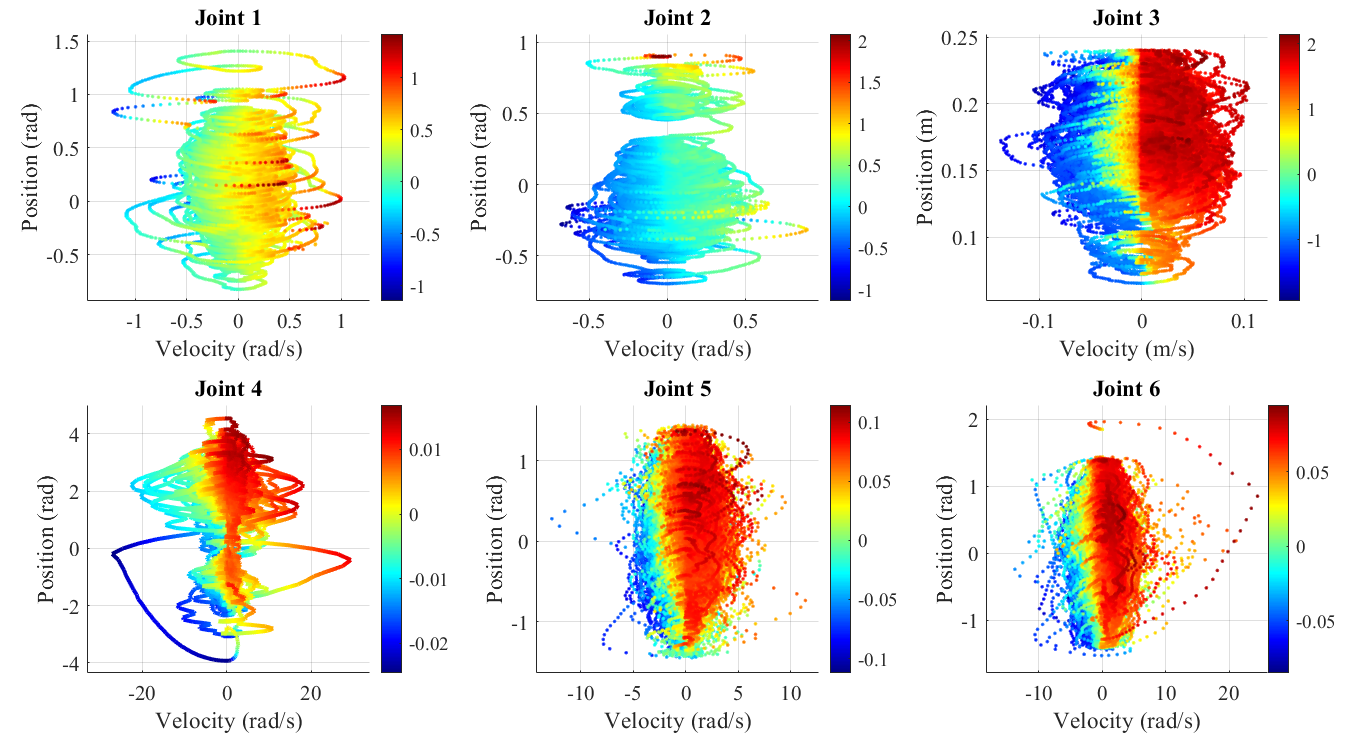}	&
\includegraphics[width=0.49\textwidth]{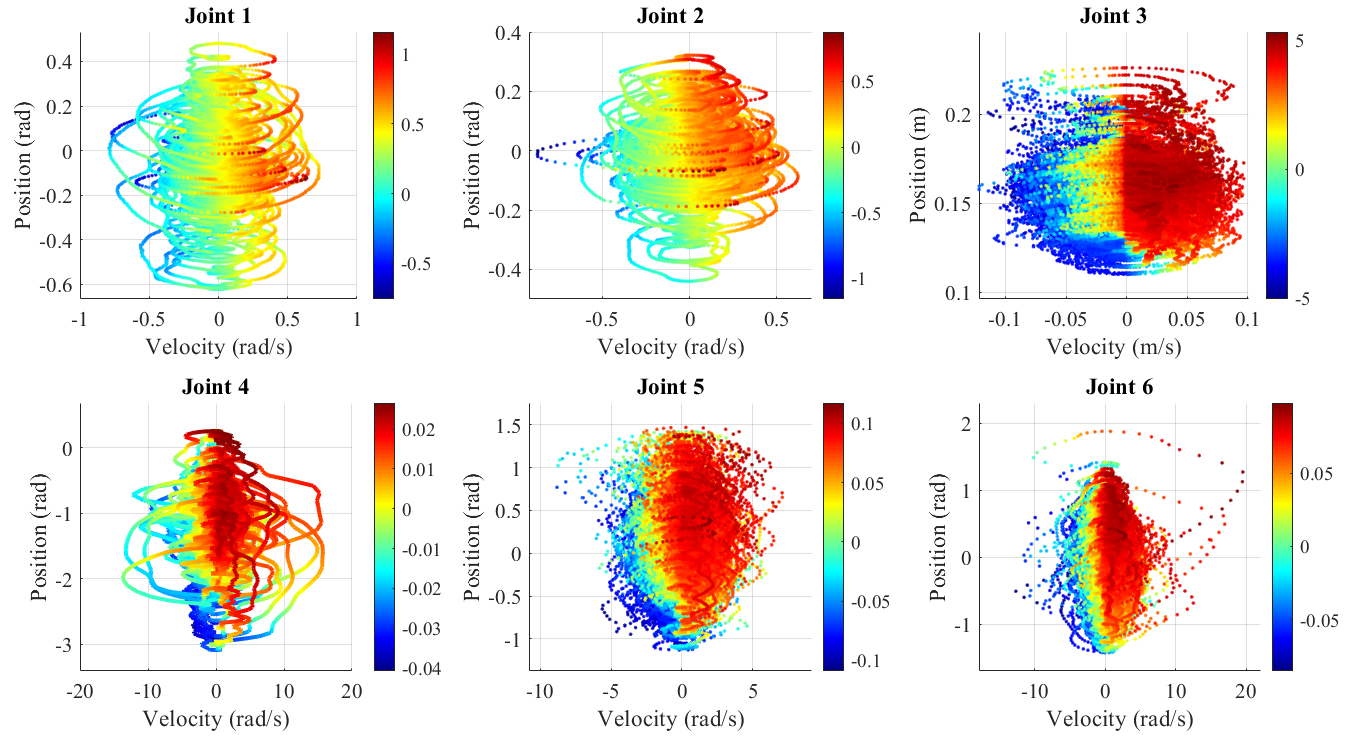} \\
(a) & (b) 
\end{tabular}
\caption{Joint states and torques used in training for (a) free-space and (b) trocar}
\label{fig:workspaces}
\end{figure*}

The abdominal phantom necessarily limits the workspace of the robot. A comparison of the free space workspace and the workspace inside the phantom is shown in Fig~\ref{fig:workspaces}.

\subsection{6 Degrees of Freedom Force Estimation}
\begin{figure}[h]
\begin{tabular}{cc}
\centering
\includegraphics[height=0.165\textheight]{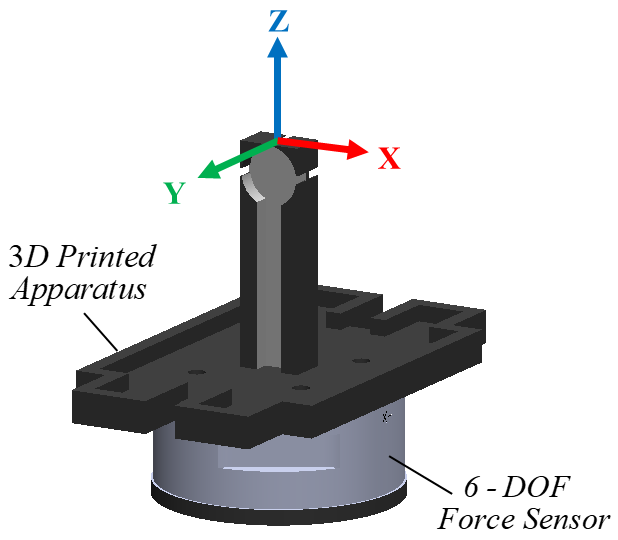}	&
\includegraphics[height=0.165\textheight]{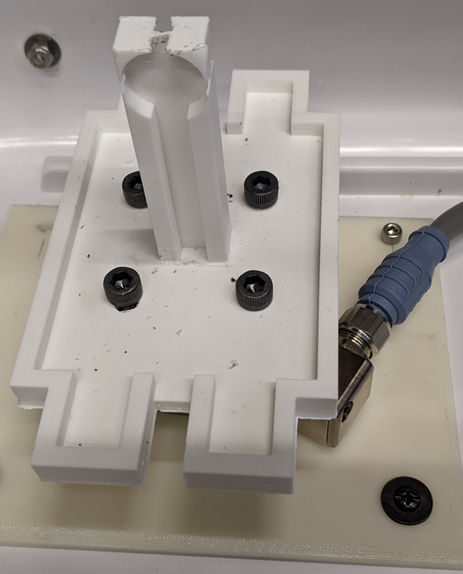} \\
\end{tabular}
\caption{CAD model (left) and test setup (right) used to collect 6 degree of freedom forces}
\label{fig:6dof_setup}
\end{figure}

We place a Gamma F/T Sensor (ATI Industrial Automation, Apex, NC, USA) in the workspace of the robot and mount a 3D printed structure on top as shown in Fig.~\ref{fig:6dof_setup}. This structure enables us to use the da Vinci instrument to grasp a handle and apply both forces and torques to evaluate 6 degrees of freedom.

\section{Results}

\subsection{Training Time vs. Training Dataset}

First, we analyze the runtime required to train each of these networks given a set of training data of a certain length. Table~\ref{table:runtime} compares the runtime to train a network from only data collected in the trocar from scratch vs. a correction network. The training was done on a Titan V (Nvidia, Santa Clara CA, USA). Since the training can be parallelized, we consider only the time it takes to train one network, for joints 5 and 6. We only analyze the time for training the network and exclude the time to read the data from disk as we assume that the small amount of data can be collected and stored in memory when actually deployed. We train \textbf{Corr} to 400 epochs. It is important to note that early stopping was only considered for the \textbf{Xfer} case as it employs a similar structure as \textbf{Troc}. The advantage should come from the transfer learning from the free space network, making the training converge faster. 

\begin{table}[h]
\centering
\setlength{\tabcolsep}{5pt}
\caption{Trocar training dataset length vs. training time. Units are [hr]:min:sec and (seconds).}
\label{table:runtime}
\begin{tabular}{|rr|rr|rr|rr|}\hline
\multicolumn{2}{|c}{\textbf{Training dataset}} & \multicolumn{2}{|c}{\textbf{Troc}} & 
\multicolumn{2}{|c}{\textbf{Corr}} & \multicolumn{2}{|c|}{\textbf{Xfer}} \\ \hline
\textbf{3:49} & (229) & 1:06:15 & (3975) & 0:39 & (39) & 0:54 & (54) \\\hline
\textbf{7:54} & (474) & 2:11:30 & (7890) & 0:44 & (44) & 1:43 & (103) \\\hline
\textbf{14:34} & (874) & 3:06:02 & (11162) & 1:09 & (69) & 2:54 & (174) \\\hline
\textbf{16:53} & (1013) & 3:58:35 & (14315) & 1:16 & (76) & 3:17 & (197) \\\hline
\end{tabular}
\end{table}

We see that both the correction network and the transfer learning approaches scale well to the size of the training set. The \textbf{Corr} network, however, scales even better than the \textbf{Xfer} network, as it does not employ an LSTM layer in the second step. While the LSTM is necessary to capture the long-term dependencies caused by the cannula seal and trocar interactions, the training time clearly shows an advantage if we pretrain it without patient-specific data to reduce training time at deployment. In contrast, the directly trained trocar network's training time increases super-linearly and, at more than an hour, is not suitable for intraoperative use. 

\subsection{No-contact Case}

As one of the key problems with not accounting for trocar forces is that they contribute falsely to forces in free-space motion, we consider the no-contact case first. The results of no contact experiments performed in the trocar can be seen in Table \ref{table:rms-joint}. These results show the fictitious forces estimated by the networks while moving freely in the phantom without any contact. It can be seen that the two step networks, \textbf{Corr} and \textbf{Xfer}, provide improvements in these errors. 

\begin{table}[h]
\centering
\caption{RMSE mean and (standard deviation) of Cartesian force ($f$, in N) and torque ($\tau$, in Nm) in free-motion over 92 trials. Networks that were not trained with the trocar show more error in no-contact motion as they do not account for cannula interaction.}
\label{table:rms-joint}
\begin{tabular}{|c|m{0.6cm}|m{0.6cm}|m{0.6cm}|m{0.6cm}|m{0.6cm}|m{0.6cm}|}\hline
\textbf{Method} & \textbf{$F_x$} &\textbf{$F_y$}  & \textbf{$F_z$}& \textbf{$\tau_x$} & \textbf{$\tau_y$} & \textbf{$\tau_z$}  \\ \hline\hline
\textbf{Troc}  & 0.80 (.26) & 0.75 (.26) &  2.02 (.66) & 0.041 (.014) & 0.039 (.013) & 0.006 (.002) \\\hline
\textbf{Base}  & 1.83 (.49) & 1.64 (.49) & 3.20 (.52) & 0.101 (.026) & 0.092 (.024) & 0.014 (.005) \\\hline
\textbf{Seal}  & 1.01 (.49) & 0.97 (.41) & 1.43 (.41) & 0.052 (.021)  & 0.051 (.020) & 0.007 (.004)\\\hline
\textbf{Base + Corr}  & 0.83 (.27) & 0.78 (.26) & 2.04 (.53) & 0.041 (.013) & 0.040 (.013) & 0.009 (.004) \\\hline
\textbf{Seal + Corr}  & 0.74 (.24) & 0.66 (.23)  & 1.16 (.56) & 0.035 (.012) & 0.037 (.011) & 0.008 (.003)\\\hline
\textbf{Seal + Xfer} & 0.47 (.11) & 0.44 (.12) & 1.08 (.39) & 0.026 (.016)  & 0.062 (.013) & 0.011 (.004) \\\hline
\end{tabular}
\end{table}

\subsection{Force Estimation}

\begin{figure*}[t]
\centering
\includegraphics[width=0.89\textwidth]{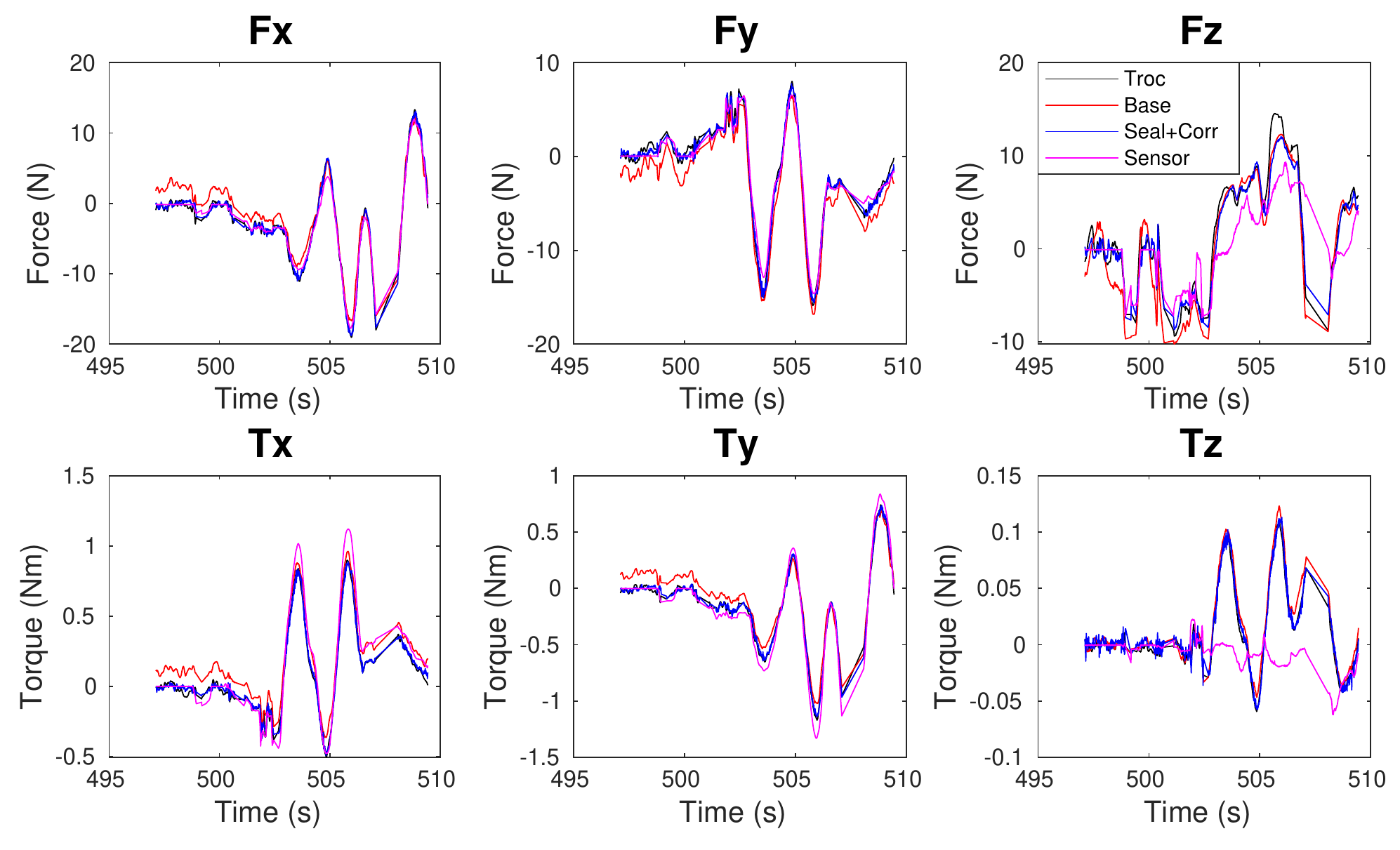}
\caption{Cartesian forces and torques for directly-trained trocar network (\textbf{Troc}), the \textbf{Base} method proposed in \cite{yilmaz2020neural}, the corrected torque when the free space network was trained with a cannula seal (\textbf{Seal+Corr}), and the force sensor measurements.}
\label{fig:force_plots}
\end{figure*}

Lastly, we verify that the proposed two step networks estimate forces more accurately inside the trocar, by interacting with the force sensor and comparing it to our network predictions. Fig.~\ref{fig:force_plots} show a sample of selected methods compared to the force sensor measurements. 

\begin{figure}[h]
\centering
\includegraphics[width=0.45\textwidth]{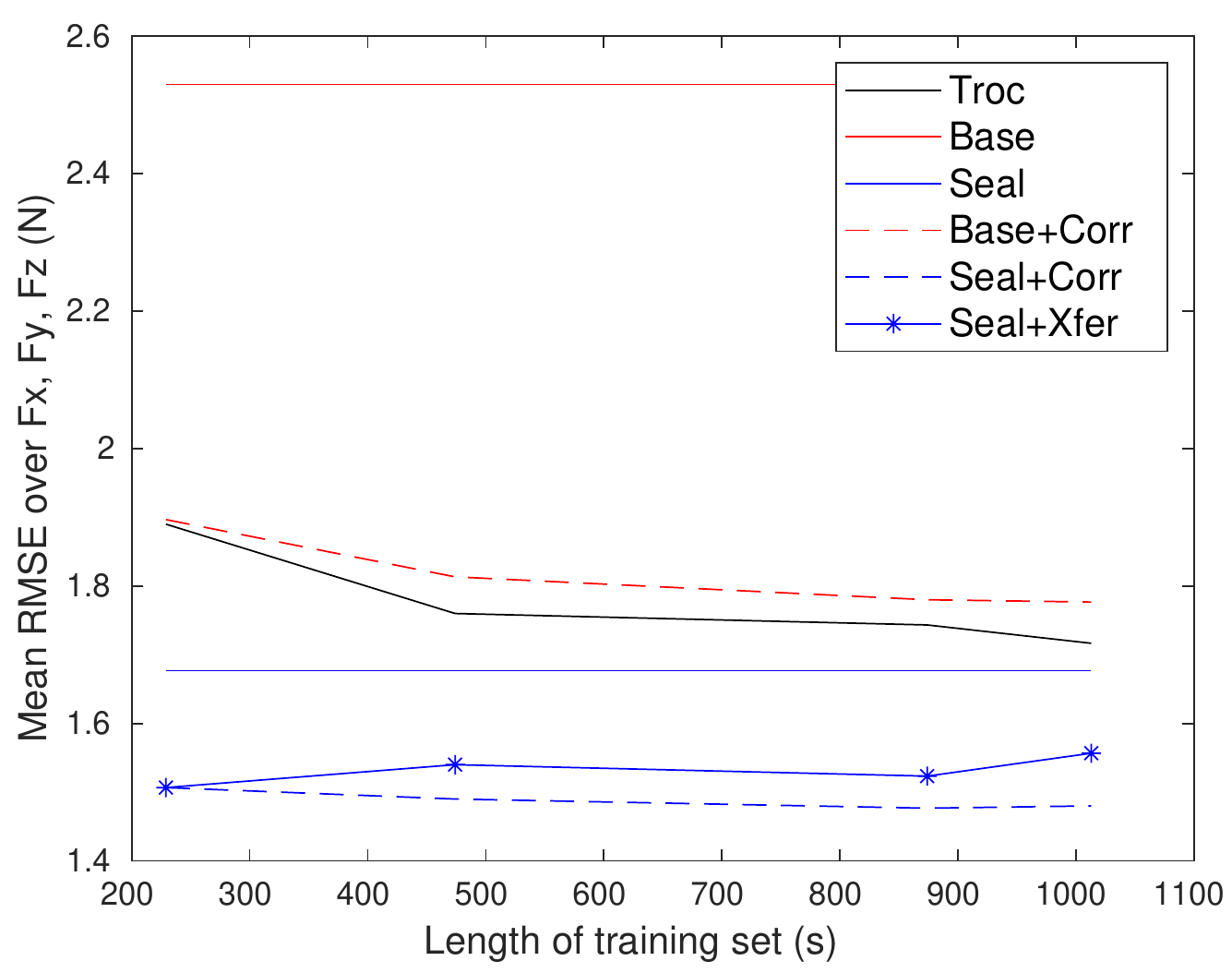}
\caption{Training set length vs. mean error (RMSE) of estimated forces over Fx, Fy, and Fz.}
\label{fig:acc_vs_time}
\end{figure}

The training set length for the plots in Fig.~\ref{fig:force_plots} is 7 min 54 s. We evaluate the RMSE over the length of the training dataset used to train each method. Fig.~\ref{fig:acc_vs_time} shows the accuracy in force estimation to illustrate how the accuracy of the RMSE changes as a function of dataset length while Table~\ref{table:rms-Cartesian} shows the full set of results. 

It can be seen that the two step approaches provide improvements in the Cartesian force estimation errors over the \textbf{Base}, \textbf{Seal} and \textbf{Troc} approaches. However, in the Cartesian torque axes, they do not show significant improvements, except over \textbf{Base}, which may be due to the wrist axes being less affected by trocar interactions. 

Table~\ref{table:rms-Cartesian} shows the RMSE versus the length of the trocar training dataset for all axes and network structures (\textbf{Base} and \textbf{Seal} are pretrained and thus do not use the trocar training data). It can be seen that the correction network and transfer learning approaches both have lower mean than the other approaches over the entire training data range, with the correction network performing slightly better. It is interesting to note that while the error of \textbf{Troc} and \textbf{Base+Corr} reduces with more training data as expected, the errors of \textbf{Seal+Corr} only reduce slightly and the errors of \textbf{Seal+Xfer} do not reduce. This could be because \textbf{Seal} already accounts for much of the error from the cannula seal. Additionally, the large learning rate used in \textbf{Xfer} speeds up training but may prevent it from converging. Furthermore, this data suggests that the training set length does not have a great significance for these networks after 200 seconds. This would imply that in a clinical scenario, data collection time can be considerably reduced. 

\begin{table}[h]
\centering
\setlength{\tabcolsep}{3pt}
\caption{RMSE mean and (standard deviation) of Cartesian force ($F$, in N) or torque ($\tau$, in Nm) for the different networks when interacting with the phantom through the trocar (12 trials for each network).
}
\label{table:rms-Cartesian}
\begin{tabular}{|c|c|r|r|r|r|}\hline
&\textbf{Method}  & 229 s & 474 s & 874 s & 1013 s \\ \hline
&\textbf{Troc}  & 1.64 (1.47)  & 1.58 (1.54) & 1.54 (1.54) & 1.53 (1.53)  \\\cline{2-6}
&\textbf{Base}  & \multicolumn{4}{c|}{2.45 (1.10)} \\\cline{2-6}
\textbf{$F_x$}&\textbf{Seal} & \multicolumn{4}{c|}{1.65 (1.53)} \\\cline{2-6} 
&\textbf{Base + Corr} & 1.60 (1.49) & 1.51 (1.43)& 1.54 (1.52) & 1.58 (1.57) \\\cline{2-6}
&\textbf{Seal + Corr} & 1.48 (1.55) & 1.42 (1.51)& 1.45 (1.54) & 1.44 (1.55) \\\cline{2-6}
&\textbf{Seal + Xfer} & 1.22 (1.49) & 1.23 (1.45) & 1.23 (1.43)  & 1.19 (1.40)   \\\hline \hline
&\textbf{Troc}  & 1.20 (0.64) & 1.16 (0.52)& 1.15 (0.52) & 1.11 (0.41) \\\cline{2-6}
&\textbf{Base}  & \multicolumn{4}{c|}{2.11 (0.51)} \\\cline{2-6}
\textbf{$F_y$}&\textbf{Seal} & \multicolumn{4}{c|}{1.26 (0.59)} \\\cline{2-6}
&\textbf{Base + Corr} & 1.28 (0.87) & 1.24 (0.81)& 1.16 (0.48) & 1.13 (0.41)\\\cline{2-6}
&\textbf{Seal + Corr} & 1.06 (0.64) & 1.03 (0.56) & 1.05 (0.48) & 1.05 (0.46)\\\cline{2-6}
&\textbf{Seal + Xfer} & 0.86 (0.43) & 0.98 (0.45)  & 0.85 (0.37)  & 1.01 (0.50)  \\\hline\hline
&\textbf{Troc}  & 2.83 (1.11) & 2.54 (0.73)& 2.54 (0.75) & 2.51 (0.72)\\\cline{2-6}
&\textbf{Base}  & \multicolumn{4}{c|}{3.03 (0.47)} \\\cline{2-6}
\textbf{$F_z$}&\textbf{Seal} & \multicolumn{4}{c|}{2.12 (0.79)} \\\cline{2-6}
&\textbf{Base + Corr} & 2.81 (0.76) & 2.69 (0.73)& 2.64 (0.76) & 2.62 (0.77)\\\cline{2-6}
&\textbf{Seal + Corr} & 1.98 (0.88) & 2.02 (0.89) & 1.93 (0.87) & 1.95 (0.89)\\\cline{2-6}
&\textbf{Seal + Xfer} & 2.44 (1.13) & 2.41 (1.05)  & 2.49 (1.21)  & 2.47 (1.10)  \\\hline \hline
&\textbf{Troc}  & .064 (.018) & .062 (.015) & .062 (.016) & .060 (.016) \\\cline{2-6}
&\textbf{Base}  & \multicolumn{4}{c|}{.122 (.025)} \\\cline{2-6}
\textbf{$\tau_x$}&\textbf{Seal} & \multicolumn{4}{c|}{.069 (.022)} \\\cline{2-6} 
&\textbf{Base + Corr} & .067 (.020)& .064 (.019)& .064 (.021) & .063 (.020)\\\cline{2-6}
&\textbf{Seal + Corr} & .061 (.021)& .060 (.021) & .059 (.019) & .060 (.022) \\\cline{2-6}
&\textbf{Seal + Xfer} & .063 (.027)  & .063 (.023)  & .060 (.024)  & .064 (.024)  \\\hline\hline
&\textbf{Troc}  & .087 (0.08) & .087 (.078)& .085 (.079) & .085 (.079) \\\cline{2-6}
&\textbf{Base}  & \multicolumn{4}{c|}{.136 (.056)} \\\cline{2-6}
\textbf{$\tau_y$}&\textbf{Seal} & \multicolumn{4}{c|}{.090 (.076)} \\\cline{2-6}
&\textbf{Base + Corr} & .087 (.080) & .084 (.076)& .084 (.077) & .087 (.080) \\\cline{2-6}
&\textbf{Seal + Corr} & .084 (.078) & .082 (.078)& .082 (.080) & .082 (.080)\\\cline{2-6}
&\textbf{Seal + Xfer} & .082 (.077)   & .080 (.075)   & .080 (.075)  & .082 (.072)  \\\hline \hline
&\textbf{Troc}  & .028 (.018) & .027 (.018)& .031 (.019) & .027 (.019)\\\cline{2-6}
&\textbf{Base}  & \multicolumn{4}{c|}{.031 (.019)} \\\cline{2-6}
\textbf{$\tau_z$}&\textbf{Seal} & \multicolumn{4}{c|}{.028 (.019)} \\\cline{2-6}
&\textbf{Base + Corr} & .029 (.017) & .031 (.018)& .027 (.019) & .027 (.018)\\\cline{2-6}
&\textbf{Seal + Corr} & .028 (.017)& .028 (.018)& .027 (.018) & .027 (.019)\\\cline{2-6}
&\textbf{Seal + Xfer} & .029 (.200)  & .030 (.020)  & .030 (.020) & .030 (.020)  \\\hline

\end{tabular}
\end{table}

The experiment results show that both two step network approaches presented provide improvements in force estimation errors in free motion and contact. With their current implementations, each method has its advantages. The two step correction network is the fastest to train and the transfer learning method has better error rates in free motion. In contact force estimation, the correction network does better in $F_z$, whereas the other directions are comparable. 

\section{Discussion}

In our experimental setup, we observed that the trocar interaction forces on the phantom abdomen are low and the cannula seal is the largest contributor to our errors and therefore also provides the best opportunity for improvement. Indeed, a network trained with data collected outside the trocar, but with a cannula seal, does fairly well at estimating forces.

Unlike our phantom, patient body musculature and fat can show greater resistance and cause larger interaction forces at the trocar. Furthermore, the patient abdominal cavities are insufflated at the beginning of the operation and then the insufflation level changes during the procedure. These factors can cause changes in the interaction forces during the operation. Since the proposed method is self-supervised, it has the potential for lifelong learning to adjust for changes in insufflation levels. If the measured torques differ slightly from predicted, we can assume the motion is in free space and use it to update the network. Otherwise, if the measured torques differ greatly from the predicted, the change is more likely caused by contact. 

Another limitation of this work is that due to equipment limitation, we used the same cannula seal for training and testing. This gives an advantage to the \textbf{Seal} setup as the third joint is generally responsible for much of the applied force and is not affected by the trocar. However, we do demonstrate that in the absence of the cannula seal, our correction and transfer learning approaches were able to improve the performance. This indicates that it could correct for changes in the cannula seal and account for more significant body wall forces in a patient as well.
All correction and direct training networks showed little improvement for $\tau_z$, but the applied torque range, and thus the absolute torque estimation error, were relatively small for that component.

The present experiments were performed with a single da Vinci instrument. Because the system can detect which instrument is installed, it is feasible to have a different network trained for each instrument type. However, it is likely that different instruments of the same type would behave differently and even the same instrument could change due to the cleaning and sterilization procedures. Thus, we anticipate using an online correction network to adapt to the instrument, similar to the adaptation to the cannula seal and trocar.

\section{Conclusion}
Previous work demonstrated that neural networks can predict internal torques of the da Vinci PSM, thereby enabling estimation of tool-to-tissue interaction forces, at least in a laboratory setting. But, this does not necessarily translate to clinical conditions, which are subject to additional external forces, due to the cannula seal and trocar, that can vary significantly based on the patient and clinical setup. It is not practical, however, to collect a large amount of training data or to wait for the network to complete training prior to starting the surgical procedure.
Therefore, in this work, we propose self-supervised, two-step training methods to account for patient-specific interactions. The proposed schemes use a network pretrained on large amounts of free space data as the backbone. We evaluate two schemes that can be quickly trained and deployed in the operating room. With 4 min of patient-specific data or less, and 1 min to train a second-step network, we estimate forces inside a phantom that are similar to those estimated in free space~\cite{tran2020}. This suggests that the proposed method may be feasible to deploy in clinical use.

\section{Acknowledgements}
This work was supported in part by NSF OISE 1927354.
The Titan V used for this research was donated by the Nvidia Corporation.

\bibliographystyle{IEEEtran}
\bibliography{references}

\end{document}